\documentclass[letterpaper, 10 pt, conference]{ieeeconf}  

\IEEEoverridecommandlockouts                              

\usepackage{graphics} 
\usepackage{epsfig} 
\usepackage{mathptmx} 
\usepackage{times} 
\usepackage{amsmath} 
\usepackage{amssymb}  
\usepackage{empheq}
\usepackage{url}
\usepackage{mathrsfs}
\usepackage{overpic}
\usepackage{adjustbox}
\usepackage{subfigure}
\usepackage{stfloats}
\usepackage{float}

\newcommand{\refeqn}[1]{(\ref{eqn:#1})}
\renewcommand{\Re}   {\ensuremath{\mathbb{R}}}

\newcommand{\SO}     {\ensuremath{\mathsf{SO(3)}}}
\newcommand{\T}      {\ensuremath{\mathsf{T}}}

\graphicspath{{./Figures/}}

\title{\LARGE \bf
Force-Motion Control For A Six Degree-Of-Freedom Robotic Manipulator
}

\author{Sagar Ojha, Karl Leodler, Lou Barbieri,  and TseHuai Wu$^{*}$
\thanks{*This research has been supported in part by Maryland Industrial Partnerships
	Program.}
\thanks{Sagar Ojha and Tse-Huai Wu, Mechanical Engineering, University of Maryland Baltimore County, Baltimore, MD 21250
        {\tt\small as03050, tsewu@umbc.edu }}%
\thanks{Karl Leodler and Lou Barbieri, Dynamic Dimension Technologies LLC, Westminster, MD 21158
        {\tt\small kleodler, lbarbier@dynamicdimensiontechnologies.com }}%
}

\begin{document}

\maketitle
\thispagestyle{empty}
\pagestyle{empty}

\begin{abstract}

This paper presents a unified algorithm for motion and force control for a six degree-of-freedom spatial manipulator. The motion-force contoller performs trajectory tracking, maneuvering the manipulator’s end-effector through desired positions, orientations and rates. When contacting an obstacle or target object, the force module of the controller restricts the manipulator movements with a novel force exertion method, which prevents damage to the manipulator, end-effectors and objects during the contact or collision. The core strategy presented in this paper is to design the linear acceleration for the end-effector which ensures both trajectory tracking and restriction of any contact force at the end-effector. The design of the controller has been validated through numerical simulations and digital twin visualization.

\end{abstract}

\section{INTRODUCTION}

Robotic manipulators are used in various industries such as automotive and aerospace for a vast amount of applications. These common applications, such as material handling and assembly, require the end effector to follow the reference trajectories. In addition to trajectory tracking, a safe collaborative robot must control the force that the end-effector exerts upon contact with any obstacles during trajectory tracking. Specifically, the magnitude of the force that the robot exerts should be bounded by the maximum allowable force.

A classical example where both position tracking and force control are required is on a robot manipulator with an eraser mounted at the end-effector to erase a whiteboard~\cite{lynch}. The manipulator pushes the eraser towards the board while moving the eraser on the plane of the board, thereby performing trajectory tracking about the plane of the board and controlling the force towards the board.

In general, motion controllers can be categorized by set-point tracking (also known as stabilization) and trajectory tracking. The proportional–integral–derivative (PID) controllers are commonly used for set-point tracking, i.e., tracking a constant desired value~\cite{siciliano}. One of the simple designs to acheive set-point tracking is a PD control with a gravity compensation algorithm~\cite{takegaki}. PID controllers are adequate for set-point tracking, however, they are ineffective for trajectory tracking, i.e., tracking a time-dependent desired value. Therefore, other techniques such as inverse dynamics control, feedback linearization technique, and passivity-based control method are utilized for effective trajectory tracking~\cite{springerhandbook},~\cite{textbook475}.

The contact force could be controlled indirectly using impedance or admittance control strategies~\cite{springerhandbook}. The indirect methods use modifications to the motion control algorithm to regulate the contact force. Indirect force controller could also be designed in a different approach than using the impedance control strategies. One example of the indirect force controller is designed by manipulating the desired trajectory in order to regulate force~\cite{xie}. The direct force control could be performed through the closure of the force feedback loop~\cite{springerhandbook}. A hybrid force-impedance controller which could perform accurate force tracking in constraint directions and compliant behavior in free motion direction is also available~\cite{iskandar}, and such controller could be interpreted as a mix of direct and indirect force control strategies.

Tasks such as polishing a surface require both motion and force control. Historically, in order to acheive trajectory tracking and force control, \textit{hybrid controllers} have been developed -- motion and force controllers are designed separately and then combined together. The hybrid controller explicitly performs trajectory tracking in certain directions and force regulation in others. In other words, the controller makes a binary decision to either perform motion control or force control for each direction. The fact that the motion and force are controlled in different directions implies that the hybrid controller can be decoupled into motion and force controllers. Moreover, the concept of the aforementioned hybrid controller can be extended to design a more general type of hybrid controller in order to perform trajectory tracking and force control in the same direction~\cite{raibert, capisani, sakaino, khatib}. The controllers sum the individual outputs of the position controller and the force controller and feeds the total output to the manipulator. In the architecture of the general hybrid controller, the motion and force controllers are linked at a distinct location where the signals of each controllers are summed. Since the motion and force controllers are linked at a distinct location, the hybrid controller can be decoupled into the motion and the force controllers as well.

The objective of the hybrid controllers is to follow the trajectory when there are not any obstacles in the trajectory and exert limited amount of force onto the obstacle when the obstacle obstructs the trajectory. More specifically, the hybrid controller performs motion control while the trajectory is in the free space, and performs force control while the trajectory is in the obstacle space. The end-effector is said to be in \textit{free space} when it is not in contact with obstacle and it is in \textit{obstacle space} when it encounters obstacles.

Unlike the aforementioned hybrid controllers where motion controller and force controller are designed separately, there are \textit{combined controllers} that are designed by integrating motion and force controller into a single design comprehensively~\cite{xu, tarikieej, tarikieee}. As such, the combined controller cannot be explicitly separated into motion and force controllers. Specifically, for the combined controller, a unified algorithm that meets the objectives of both motion and force controllers is designed in a holistic approach.

The combined controller also performs trajectory tracking and force control in the same direction; however, the design of the combined controller is different than the hybrid controller. The combined controller transitions between motion control and force control modes depending on whether the end-effector is in free space or the obstacle space.

Inspired from~\cite{tarikieej} and~\cite{tarikieee}, a combined motion-force controller that performs motion and force control in the same direction is developed for the 6 DOF manipulator. Although the similar idea has been implemented on 1 DOF and 2 DOF planar robots before, in-depth analysis of the controller is missing in the works presented in the literature. 

This paper develops and implements a combined controller for a six degree-of-freedom spatial manipulator which poses a significant challenge in the derivation of the governing equation of motion compared to the planar robots. Further, providing an in-depth analysis for the presented  motion-force controller design is another contribution. The paper analyzes the dynamics of the robot and presents the model for contact force between the end-effector and the environment in Section II. The motion and force control strategy is presented in Section III. Numerical simulations and a brief discussion of the results are presented in Section IV followed by the conclusions in Section V.

\section{PRELIMINARIES}
\subsection{Dynamics Analysis for the 6 DOF Manipulator}

One can obtain the equations of motion for the 6 DOF manipulator using Lagrangian method, which is a common way to derive the Euler-Lagrange equations of motion for a multi rigid-body system~\cite{textbook475}. The Lagrangian $L$ of a system is defined as
\begin{align}
	L \triangleq K.E. - P.E.,	\label{eqn:Lagrangedefinition}
\end{align}
where $K.E.$, $P.E. \in\Re$ are the total kinetic and potential energies of the multi rigid-body system, respectively.

The kinetic energy $\kappa$ of a rigid body, is given by
\begin{align}
	\kappa = \frac{1}{2} {m} {v^\T} {v} + \frac{1}{2} {\omega}^\T {\mathcal{I}} {\omega}, \label{eqn:kinetic_energy_1_body}
\end{align}
where $m \in\Re^{+}$ is the mass of the body. $v$, $\omega \in\Re^3$ are the linear and angular velocities of the body expressed in inertial frame, and $\mathcal{I} \in\Re^{3 \times 3}$ is the matrix representing the inertia of the body expressed with respect to the inertial reference frame. The inertia matrix $\mathcal{I}$ is obtained as
\begin{align}
	\mathcal{I} = R I R^{\T}, \label{eqn:inertial_I}
\end{align}
where $R \in\SO$ is the rotation matrix representing the orientation of the body attached coordinate frame expressed with respect to the inertial coordinate frame, and $I \in\Re^{3 \times 3}$ is the inertia matrix of the rigid-body expressed with respect to body attached coordinate frame.

\begin{figure}[h]
	\centerline{
		\includegraphics[width=0.39\columnwidth]{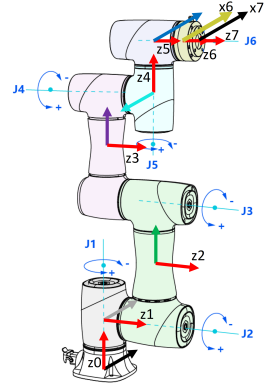}
	}	
	\caption{Coordinate frame assignment for CR3 showing x and z axes of all the coordinate frames that are assigned to the robot. The coordinate frames are assigned at the center of mass of each link. Note that the bottom piece is fixed to the ground; hence, it does not influence the dynamics of the manipulator. Also, frame {7} has been assigned at the end-effector location. (Original image credit: Dobot CR3 User Guide)}\label{fig:cogframes}
\end{figure}

The body attached coordinate frames for each link of the 6 DOF manipulator are shown in Fig.~\ref{fig:cogframes}. Note that the coordinate frame assignment shown in Fig.~\ref{fig:cogframes} is for dynamics analysis, which is different from the coordinate frame assignment for kinematic analysis~\cite{ourpaper}. In particular, the paper adapts the Denavit-Hartenberg (DH) convention for kinematic analysis.

The total kinetic energy $K.E.$ of the 6 DOF manipulator is derived as
\begin{align}
	K.E. = \sum_{i=1}^{6}{{\kappa}_i} &= \frac{1}{2} \sum_{i=1}^{6}{[ m_{i} v_i^\T v_i + \omega_i^{\T} \mathcal{I}_i \omega_i ]} \\
	   &= \frac{1}{2} \sum_{i=1}^{6}{[ m_{i} v_i^\T v_i + \omega_i^{\T} R_{i} I_{i} R_{i}^{\T} \omega_i ]}. \label{eqn:totalkineticenergy_multibody}
\end{align}
Let $\dot{\theta} \in\Re^{6}$ be the joint velocity, $v_i \in\Re^{3}$ be the linear velocity of the origin of coordinate frame $\{i\}$ expressed with respect to frame $\{0\}$, and $\omega_i \in\Re^{3}$ be the angular velocity of the coordinate frame $\{i\}$ expressed with respect to frame $\{0\}$. Let $J_i \in\Re^{6 \times 6} \text{ for } i=1,2,\ldots,7$ be the Jacobian matrix expressed with respect to the base frame $\{0\}$ that represents the linear transformation from $\dot{\theta}$ to $v_i$ and $\omega_i$  as shown in Eq.~\refeqn{Jacobian_for_frames},
\begin{align}
	\begin{bmatrix}
		v_i \\
		\omega_i
	\end{bmatrix}
	&=
	J_i
	\begin{bmatrix}
		\dot{\theta}_1 \\
		\vdots \\
		\dot{\theta}_6
	\end{bmatrix}
	=
	\begin{bmatrix}
		J_{v_i} \\
		J_{\omega_i}
	\end{bmatrix}
	\begin{bmatrix}
		\dot{\theta}_1 \\
		\vdots \\
		\dot{\theta}_6
	\end{bmatrix},
	\label{eqn:Jacobian_for_frames}
\end{align}
where $J_{v_i}, J_{\omega_i} \in\Re^{3 \times 6}$ are the upper and lower halves of $J_i$.

The linear and angular velocities of frame $\{i\}$ could thus be obtained from the joint velocity as
\begin{align}
	v_i = J_{v_i} \dot{\theta},\quad \omega_i = J_{\omega_i} \dot{\theta}. \label{eqn:joint2taskvel}
\end{align}

Substituting Eq.~\refeqn{joint2taskvel} to Eq.~\refeqn{totalkineticenergy_multibody} gives
\begin{align}
	K.E. &= \frac{1}{2} \sum_{i=1}^{6}{[ m_{i} (J_{v_i} \dot{\theta})^{\T} (J_{v_i} \dot{\theta}) + (J_{\omega_i} \dot{\theta})^{\T} R_{i} I_{i} R_{i}^{\T} (J_{\omega_i} \dot{\theta}) ]}, \\
	   &= \frac{1}{2} \sum_{i=1}^{6}{[ m_{i} \dot{\theta}^{\T} J_{v_i}^{\T} J_{v_i} \dot{\theta} + \dot{\theta}^{\T} J_{\omega_i}^{\T} R_{i} I_{i} R_{i}^{\T} J_{\omega_i} \dot{\theta} ]}, \\
	   &= \frac{1}{2} \sum_{i=1}^{6}{({\dot{\theta}^{\T}}[ m_{i} J_{v_i}^{\T} J_{v_i} + J_{\omega_i}^{\T} R_{i} I_{i} R_{i}^{\T} J_{\omega_i}}] \dot{\theta}), \\
	   &= \frac{1}{2} {\dot{\theta}^{\T} \sum_{i=1}^{6} [ m_{i} J_{v_i}^{\T} J_{v_i} + J_{\omega_i}^{\T} R_{i} I_{i} R_{i}^{\T} J_{\omega_i}} ] \; \dot{\theta}, \\
   	&= \frac{1}{2} \dot{\theta}^{\T} A \; \dot{\theta}, \label{eqn:totalkineticenergy_2}
\end{align}
where $A \in\Re^{6 \times 6}$ is the mass matrix of the 6 DOF manipulator.

Substituting Eq.~\refeqn{totalkineticenergy_2} to Eq.~\refeqn{Lagrangedefinition} gives
\begin{align}
	L &= \frac{1}{2} \dot{\theta}^{\T} A \; \dot{\theta} + P.E. =  \frac{1}{2} \sum_{i, j}^{6}{ a_{ij}\dot{\theta}_{i}\dot{\theta}_{j} } + P.E., \label{eqn:Lagrangesubstitution}
\end{align}
where $a_{ij}$ is the element at the $i$-{th} row and $j$-{th} column of $A$.

The Euler-Lagrange equation of motion is given as
\begin{align}
	\frac{\partial}{\partial{t}} \frac{\partial{L}}{\partial{\dot{\theta}_{k}}} - \frac{\partial{L}}{\partial{\theta}_{k}} = \tau_{k}  \text{ where } k=1,2,\ldots,6. \label{eqn:EulerLagrange1}
\end{align}
Here, $\tau_{k}$ is the $k$-th joint torque. After substituting the expression for $L$ from Eq.~\refeqn{Lagrangesubstitution} to Eq.~\refeqn{EulerLagrange1} and simplifying the obtained expression, one gets
\begin{align}
	\sum_{j=1}^{6} { {a_{kj} \ddot{\theta}_{j}} } + \sum_{i, j}^{6} { \left\{ \frac{\partial{a_{kj}}}{\partial{\theta_{i}}} - \frac{1}{2} \frac{\partial{a_{ij}}}{\partial{\theta_{k}}} \right\} \dot{\theta}_{i} \dot{\theta}_{j} } - \frac{\partial{P.E.}}{\partial{\theta}_{k}} &= \tau_{k}, \label{eqn:EulerLagrange2}
\end{align}
where $k=1,2,\ldots,6$. One can use the fact that $A$ is a symmetric matrix to further simplify Eq.~\refeqn{EulerLagrange2} as
\begin{align}
	\sum_{j=1}^{6} { {a_{kj} \ddot{\theta}_{j}} } + \sum_{i, j}^{6} \frac{1}{2}{ \left\{ \frac{\partial{a_{kj}}}{\partial{\theta_{i}}} + \frac{\partial{a_{ki}}}{\partial{\theta_{j}}} - \frac{\partial{a_{ij}}}{\partial{\theta_{k}}} \right\} \dot{\theta}_{i} \dot{\theta}_{j} } - \frac{\partial{P.E.}}{\partial{\theta}_{k}} &= \tau_{k}. \label{eqn:EulerLagrange3}
\end{align}

Eq.~\refeqn{EulerLagrange3} could be expressed in matrix form as shown in Eq.~\refeqn{generaldynamics}. After the inclusion of the external forces exerted at the end-effector by the environment, the matrix-form of equation of the motion for the 6 DOF manipulator could be expressed as
\begin{align}
	\tau = A(\theta)\ddot{\theta} + B(\theta, \dot{\theta})\dot{\theta} + g(\theta) + \tau^{ext}, \label{eqn:generaldynamics}
\end{align}
where $\theta$, $\dot{\theta}$, $\ddot{\theta} \in\Re^6$ are the joint position, joint velocity, and joint acceleration of the manipulator, $A$ is the mass matrix, $B\in\Re^{6 \times 6}$ is the coriolis matrix, $g\in\Re^6$ consists of gravity forces for each link of the manipulator, $\tau\in\Re^6$ is the torque applied at each joint, and $\tau^{ext}\in\Re^6$ is the external forces and torques applied at the end-effector. Also, note that the element of $B$ at $k$-th row and $j$-th column is obtained as
\begin{align}
	b_{kj} = \sum_{i=1}^{6}{\frac{1}{2} \left\{ \frac{\partial{a_{kj}}}{\partial{\theta_{j}}} + \frac{\partial{a_{ki}}}{\partial{\theta_{j}}} - \frac{\partial{a_{ij}}}{\partial{\theta_{k}}} \right\}  \dot{\theta}_\textit{i}}. \label{eqn:B_element}
\end{align}

\subsection{Force Modeling}

\begin{figure}[h]
	\centering \includegraphics[width=0.23\textwidth]{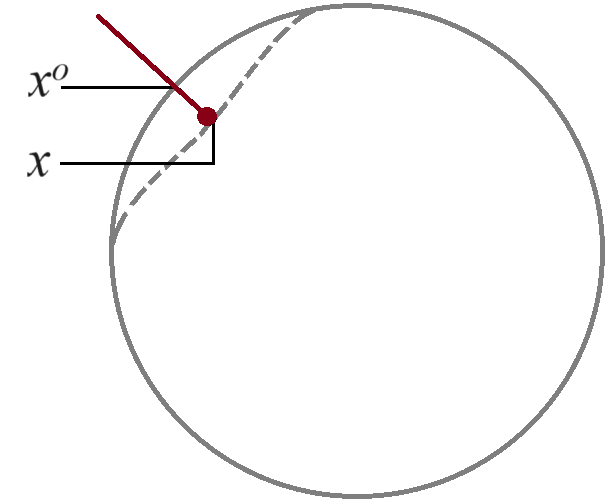}
	\caption[Spring-Damper force model]{Spring-damper force model. The original boundary of the obstacle (solid gray) is distorted (dashed gray) when the end-effector tip (red dot) hits the obstacle. $x$ is the end-effector position and $x^{o}$ is the obstacle position before the elastic deformation.}
\label{fig:force_model}
\end{figure}

The force that the end-effector exerts onto the obstacle is denoted by $f \in\Re^3$, and it is modeled as a spring-damper force, which is a common method for modeling contact force. A simple spring-damper model is shown in Fig.~\ref{fig:force_model} where the boundary of the obstacle is modeled as a spring-damper system. The amount of elastic deformation of the surface gives rise to the spring force whereas the rate change of the amount of deformation gives rise to damping force. Therefore, the force exerted by the end-effector is
\begin{align}
	f = \begin{cases}
					  K^{o}(x - x^{o}) + D^{o}(\dot{x} - \dot{x}^{o}) & \text{in obstacle-space}; \\
					  \mathbf{0} & \text{in free-space},
					  \end{cases}
	\label{eqn:force_exerted}
\end{align}
where $x \in\Re^3$ is the position of the end-effector, $x^{o} \in\Re^3$ is the position of the contact point on the boundary of the obstacle before the elastic deformation. $K^{o}$, $D^{o} \in\Re^{3 \times 3}$ are diagonal matrices whose components are the spring constants and damping coefficients of the obstacle where the end-effector makes the contact. Also, note that $\mathbf{0} \in\Re^3$.

\section{MOTION \& FORCE CONTROL}
\subsection{Controller Design Overview}

The combined controller is designed to meet the control objectives in both free space and obstacle space by integrating the motion control and force control into a single comprehensive design. Specifically, the force controller is built on top of a modified motion controller; thus, the subsequent design builds on top of the previous design. In summary, a unified algorithm that meets the objectives in both free and obstacle spaces is designed in a holistic approach.

\subsection{Controller Design without Obstacle}

The task of the combined controller in free space, i.e., when the end-effector is not in contact with an obstacle, is to make the end-effector follow the given trajectory. The controller acheives trajectory tracking by computing the joint torques for the 6 DOF manipulator that minimizes the position error between the end-effector and the trajectory point. The position error $e \in\Re^3$ is defined as
\begin{align}
	e \triangleq x - x^{ref}, \label{eqn:positionerror}
\end{align}
where $x$, $x^{ref} \in\Re^3$ are the position vectors of the end-effector and trajectory expressed with respect to frame $\{0\}$.

It is a common approach to design a motion controller based on the dynamics of the position error $e$. While the end-effector is in free space, the presented combined controller seeks to drive the position error $e$ to $\mathbf{0}$. The dynamics of the position error $e$ is modeled as
\begin{align}
	\dot{e} + Ce = \bar{e}, \label{eqn:positionerrordynamics}
\end{align}
where $\dot{e} \in\Re^3$ is the first order derivative of $e$ with respect to time; thus, $\dot{e}$ is the velocity error. $C\in\Re^{3 \times 3}$, also known as the \textit{gain matrix}, is a diagonal matrix, and $\bar{e} \in\Re^3$ is the non-homogeneous term. In fact, $\bar{e}$ is an error based on both position error $e$ and velocity error $\dot{e}$.

Eq.~\refeqn{positionerrordynamics} could be expressed in expanded form as
\begin{align}	
	\begin{bmatrix}
		\dot{e}_x \\
		\dot{e}_y \\
		\dot{e}_z
	\end{bmatrix}
	&=
	\begin{bmatrix}
		-C_x & 0 & 0 \\
		0 & -C_y & 0 \\
		0 & 0 & -C_z
	\end{bmatrix}
	\begin{bmatrix}
		e_x \\
		e_y \\
		e_z
	\end{bmatrix}
	+
	\begin{bmatrix}
		\bar{e}_x \\
		\bar{e}_y \\
		\bar{e}_z
	\end{bmatrix} \\
	&=
	\begin{bmatrix}
		-C_x e_x + \bar{e}_x\\
		-C_y e_y + \bar{e}_y \\
		-C_z e_z + \bar{e}_z
	\end{bmatrix},
	\label{eqn:positionerrordynamicsexpand}
\end{align}
where $e_x$, $e_y$, $e_z \in\Re$ and $\dot{e}_x$, $\dot{e}_y$, $\dot{e}_z \in\Re$ are the elements of $e$ and $\dot{e}$, respectively, $\bar{e}_x$, $\bar{e}_y$, $\bar{e}_z \in\Re$ are the elements of $\bar{e}$, and $C_x$, $C_y$, $C_z \in\Re$ are the diagonal elements of $C$.

Eq.~\refeqn{positionerrordynamicsexpand} expresses a system of decoupled nonhomogeneous first order linear differential equations whose solution is as follows:
\begin{align}
	\begin{bmatrix}
		{e}_x \\
		{e}_y \\
		{e}_z
	\end{bmatrix}
	&=
	\begin{bmatrix}
		\mathbf{e}^{-C_x.t}.e_x(0) + \frac{\bar{e}_x}{C_x} ( 1 - \mathbf{e}^{-C_x.t}) \\[5pt]
		\mathbf{e}^{-C_y.t}.e_y(0) + \frac{\bar{e}_y}{C_y} ( 1 - \mathbf{e}^{-C_y.t}) \\[5pt]
		\mathbf{e}^{-C_z.t}.e_z(0) + \frac{\bar{e}_z}{C_z} ( 1 - \mathbf{e}^{-C_z.t} )
	\end{bmatrix},
	\label{eqn:positionerrordynamicssoln1}
\end{align}
where $\mathbf{e}$ refers to the exponential constant, $t$ is the time and $e_x(0)$, $e_y(0)$, and $e_z(0)$ are the initial values of $e_x$, $e_y$, and $e_z$, respectively.

Eq.~\refeqn{positionerrordynamicssoln1} shows that $\lim\limits_{t \rightarrow \infty} e = \mathbf{0}$, i.e., the end-effector converges to the reference trajectory asymptotically when the elements of $C$ are positive and
\begin{align}
	\bar{e} = \mathbf{0}. \label{eqn:sigmais0}
\end{align}
Therefore, in order to perform trajectory tracking, the control input, i.e., the joint torque $\tau$ should be selected to enforce the condition expressed in Eq.~\refeqn{sigmais0}.

\subsection{Controller Design with Obstacle}

\begin{figure}[h]
	\centerline{
		{\includegraphics[width=0.35\columnwidth]{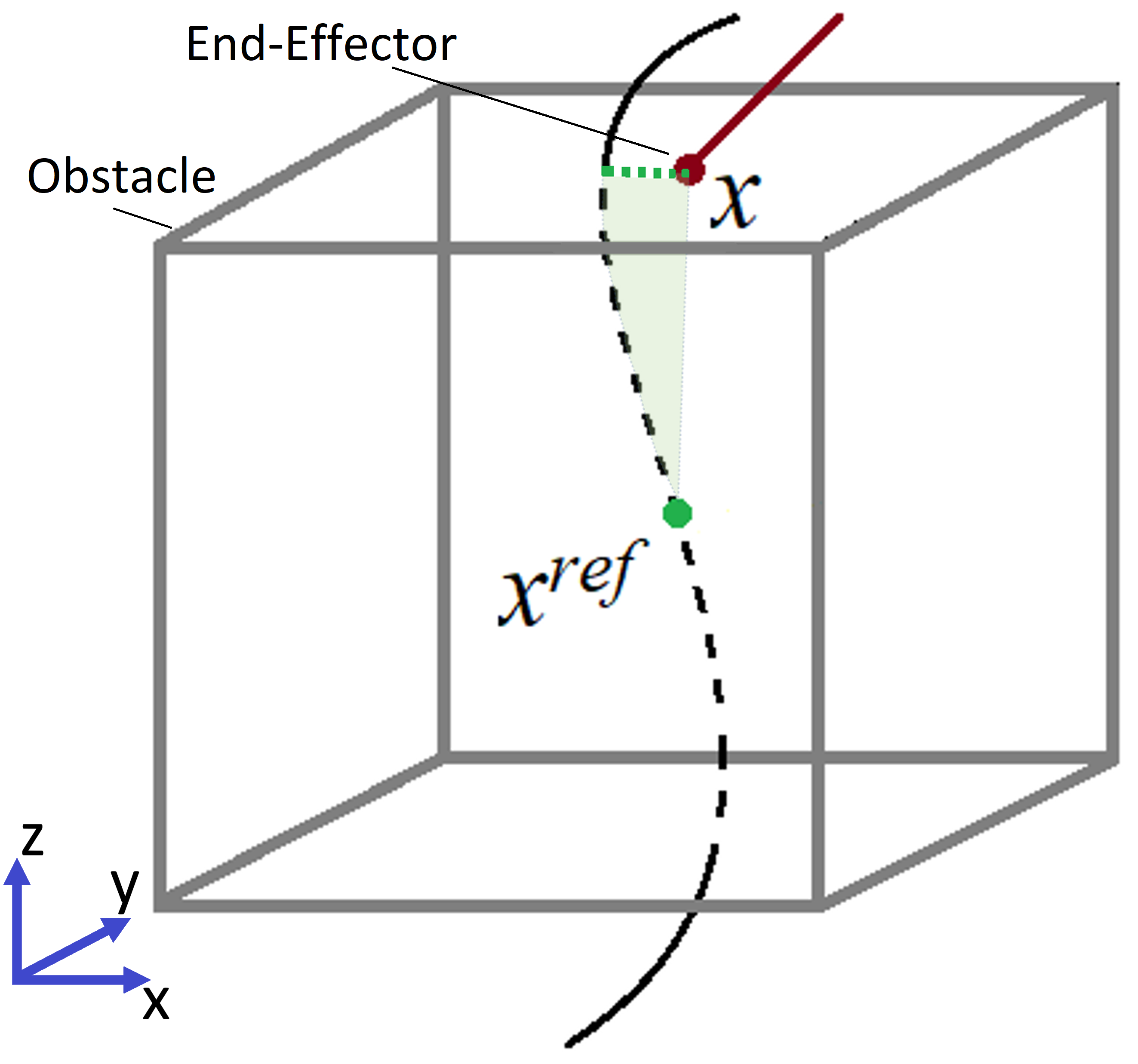}}
	}	
	\caption[Trajectory inside the obstacle space]{Solid and dotted black lines represent the reference path in the free and obstacle spaces respectively. The red colored line represents the end-effector of the manipulator and the red dot represents the tip of the end-effector. The green dotted line is the projection of the reference path onto the top surface of the cube. At a particular time, the reference trajectory $x^{ref}$ is inside the obstacle. In this case, the end-effector (red dot) could acheive trajectory tracking along $x$ and $y$ axes whereas it cannot track the trajectory along the $z$ axis. As a result, the magnitude of the $z$-component of $e$ increases.}\label{fig:trajectoryinobstacle}
\end{figure}

Suppose that the end-effector encounters an obstacle on its way while tracking the reference trajectory; the position error $e$ would start to increase as the reference trajectory $x^{ref}$ starts to get farther from the end-effector location $x$. In particular, the magnitude of $e$ increases in the direction in which the reference trajectory gets away from the end-effector as shown in Fig.~\ref{fig:trajectoryinobstacle}. As a result, the joint torque $\tau$ increases in an effort to track the trajectory. However, the increase in joint torques would only increase the force that the end-effector applies onto the obstacle. Therefore, the task of the combined controller is to change to force control mode in order to control the magnitude of the force exerted onto the obstacle such that the force does not exceed the magnitude of the reference force.

Since the end-effector exerts force as a result of increasing position error $e$, the controller changes to force control mode when the position error $e$ reaches an allowable maximum value. Also, when the reference trajectory is in the obstacle space, the condition in Eq.~\refeqn{sigmais0} is not met; as such, the position error $e$ would reach a non-zero steady state value that is directly proportional to the error $\bar{e}$ as observed from Eq.~\refeqn{positionerrordynamics}. Therefore, $\bar{e}$ has to be bounded in order to bound $e$ when the end-effector hits an obstacle. The amount by which $\bar{e}$ is bounded is chosen to be a fraction of the magnitude of reference force $f^{ref} \in\Re^3$ as shown in Eq.~\refeqn{sigmasaturated1}.

Let the bounded $\bar{e}$ be denoted by $\bar{e}^{b} \in\Re^3$; the superscript $b$ refers to ``bound". Also, let $\ast \in \{x,\;y,\;z\}$ denote the components of a vector that belongs to $\Re^3$ as well as the diagonal elements of a matrix that belongs to $\Re^{3 \times 3}$. The magnitude of the components of $\bar{e}^{b}$ are
\begin{subequations}
\begin{empheq}[left ={\|\bar{e}^b_\ast\| =\empheqlbrace}]{align}
        \|\bar{e}_\ast\|, & \:\:\:\:\:\:\:\:\text{if } \|\bar{e}_{\ast}\| \leq \rho \|f^{ref}_{\ast}\|; \label{eqn:sigmasaturated1a} \\
        \rho \|f^{ref}_{\ast}\|, & \:\:\:\:\:\:\:\:\text{if } \|\bar{e}_{\ast}\| > \rho \|f^{ref}_{\ast}\|, \label{eqn:sigmasaturated1b}
\end{empheq}
\label{eqn:sigmasaturated1}
\end{subequations}
where $\bar{e}_{\ast}$, $\bar{e}^{b}_{\ast}$, $f^{ref}_{\ast} \in\Re$ are the components of $\bar{e}$, $\bar{e}^{b}$, and $f^{ref}$ respectively, and $0 < \rho\le 1$. From Eq.~\refeqn{sigmasaturated1}, we get the components of $\bar{e}^{b}$ as
\begin{subequations}
\begin{empheq}[left ={\bar{e}^{b}_{\ast} =\empheqlbrace}]{align}
        \bar{e}_{\ast}, & \:\:\:\:\:\:\:\:\text{if } -\rho \|f^{ref}_{\ast}\| \leq \bar{e}_{\ast} \leq \rho \|f^{ref}_{\ast}\|; \label{eqn:sigmasaturated2a} \\
        \rho \|f^{ref}_{\ast}\|, & \:\:\:\:\:\:\:\:\text{if } \bar{e}_{\ast} > \rho \|f^{ref}_{\ast}\|; \label{eqn:sigmasaturated2b} \\
        -\rho \|f^{ref}_{\ast}\|, & \:\:\:\:\:\:\:\:\text{if } \bar{e}_{\ast} < -\rho \|f^{ref}_{\ast}\| \label{eqn:sigmasaturated2c}.
\end{empheq}
\label{eqn:sigmasaturated2}
\end{subequations}

In order to formulate the control law for the combined controller, the relationship between $f$ and $\bar{e}^{b}$ has to be obtained. Their relationship is proposed using Eq.~\refeqn{sigmasaturated1} which is discussed the next. The proposed relationship ensures both motion and force control objectives.

The magnitude of $f_{\ast}$ should be bounded by the magnitude of the reference force $f^{ref}_{\ast}$ during the contact with obstacle, because the task of the combined controller is to limit the force in obstacle space. Also, from Eq.~\refeqn{sigmasaturated1}, it is observed that $\|\bar{e}^{b}_{\ast}\|$ is bounded by $\rho\|f^{ref}_{\ast}\|$. Therefore, the relationship between the magnitudes of $f_{\ast}$ and $\bar{e}^{b}_{\ast}$ during the contact with the obstacle is proposed as
\begin{align}
	\|\bar{e}^{b}_{\ast}\| = \rho \|f_{\ast}\|. \label{eqn:mag_sigma_sat_is_rho_mag_f}
\end{align}

The direction of the force $f_{\ast}$ that the end-effector exerts onto the obstacle is always opposite to the direction of the position error $e_{\ast}$. Since, the directions of $\bar{e}^{b}_{\ast}$ and the position error $e_{\ast}$ are the same, the direction of $f_{\ast}$ is opposite to $\bar{e}^{b}_{\ast}$. Therefore, the complete relationship between $f_{\ast}$ and $\bar{e}^{b}_{\ast}$ during the contact with the obstacle is proposed as
\begin{align}
	\bar{e}^{b}_{\ast} + \rho f_{\ast} = 0. \label{eqn:sigma_sat_is_n_rho_f}
\end{align}
The combined controller seeks to acheive the condition in Eq.~\refeqn{sigma_sat_is_n_rho_f}. More explicitly, both motion and force control objectives are met if condition in Eq.~\refeqn{sigma_sat_is_n_rho_f} is met. The analysis of Eq.~\refeqn{sigma_sat_is_n_rho_f} in free and obstacle spaces is as follows:

\paragraph{Free Space} The end-effector does not come in contact with the obstacle; thus, $f_{\ast} = 0$. Therefore, $\bar{e}^{b}_{\ast} = 0$ and Eq.~\refeqn{sigmasaturated2a} further implies $\bar{e}_{\ast} = 0$ which is the objective of the controller in free space as expressed in Eq.~\refeqn{sigmais0}.

\paragraph{Obstacle Space} The end-effector comes in contact with the obstacle; thus $f_{\ast} = - \bar{e}^{b}_{\ast} / \rho$. Substituting for $\bar{e}^{b}_{\ast}$ from Eq.~\refeqn{sigmasaturated2} gives
\begin{subequations}
\begin{empheq}[left ={f_{\ast} =\empheqlbrace}]{align}
       -\frac{\bar{e}_{\ast}}{\rho}, & \:\:\:\:\:\:\:\:\text{if } -\rho \|f^{ref}_{\ast}\| \leq \bar{e}_{\ast} \leq \rho \|f^{ref}_{\ast}\|; \label{eqn:force2a} \\
       -\|f^{ref}_{\ast}\|, & \:\:\:\:\:\:\:\:\text{if } \bar{e}_{\ast} > \rho \|f^{ref}_{\ast}\|; \label{eqn:force2b} \\
      	 \|f^{ref}_{\ast}\|, & \:\:\:\:\:\:\:\:\text{if } \bar{e}_{\ast} < -\rho \|f^{ref}_{\ast}\| \label{eqn:force2c}.
\end{empheq}
\label{eqn:force}
\end{subequations}
Eq.~\refeqn{force} shows that the magnitude of $f_{\ast}$ is bounded by the magnitude of $f^{ref}_{\ast}$ which is the objective of the controller in obstacle space.

To summarize, the objectives in free space and obstacle space are to track the trajectory and control the force exerted by the end-effector onto the obstacle, respectively. These objectives are met simultaneously if the combined controller meets the condition expressed in Eq.~\refeqn{sigma_sat_is_n_rho_f}.

\subsection{Controller Output}
Eq.~\refeqn{sigma_sat_is_n_rho_f}  can be expressed in vector form as
\begin{align}
	\bar{e}^b + \rho f = \mathbf{0}. \label{eqn:sigma_sat_is_n_rho_f2}
\end{align}
Define the error for the control system $e^{sys} \in\Re^3$:
\begin{align}
	e^{sys} \triangleq \bar{e}^b  + \rho f. \label{eqn:sigma_t_is_sigmasat_rho_f}
\end{align}
The objective of the controller is to meet the condition in Eq.~\refeqn{sigma_sat_is_n_rho_f2}, hence the equivalent objective of the combined controller in terms of $e^{sys}$ is to drive $e^{sys}$ to $\mathbf{0}$. Furthermore, the combined controller is constructed by enforcing the dynamics of $e^{sys}$ as
\begin{align}
	\dot{e}^{sys} + K e^{sys} = \mathbf{0}, \label{eqn:sigma_t_dynamics}
\end{align}
where $\dot{e}^{sys} \in\Re^3$ is the first time derivative of $e^{sys}$ and $K \in\Re^{3 \times 3}$ is a diagonal matrix. The analysis of the solution to Eq.~\refeqn{sigma_t_dynamics} shows that $\lim\limits_{t \rightarrow \infty} e^{sys} = \mathbf{0}$ for positive elements of $K$. The matrix $K$ is also a known as s gain matrix, and the other gain matrix $C$ is introduced in Eq.~\refeqn{positionerrordynamics}. In summary, the combined controller calculates the end-effector linear acceleration based on Eq.~\refeqn{sigma_t_dynamics}. Eq.~\refeqn{sigmatdynamics2}-\refeqn{desired_acceleration_component_33} present the process of deriving the end-effector linear acceleration based on Eq.~\refeqn{sigma_t_dynamics} through the mere substitutions of the variables.

Substituting the expression for $e^{sys}$ from Eq.~\refeqn{sigma_t_is_sigmasat_rho_f} to Eq.~\refeqn{sigma_t_dynamics}, we get
\begin{align}
	\dot{\bar{e}}^{b} + \rho \dot{f} = -K e^{sys}, \label{eqn:sigmatdynamics2}
\end{align}
where $\dot{\bar{e}}^{b} \in\Re^3$ is the first time derivative of $\bar{e}$. The component form of Eq.~\refeqn{sigmatdynamics2} is
\begin{align}
	-K_{\ast} e^{sys}_{\ast} = \dot{\bar{e}}^{b}_{\ast} + \rho \dot{f}_{\ast}, \label{eqn:sigmatdynamics3}
\end{align}
where $K_{\ast} \in\Re$ is the diagonal element of the gain matrix $K$. Next, to obtain the expression of $\dot{\bar{e}}^{b}_{\ast}$ for Eq.~\refeqn{sigmatdynamics3}, one can take the first order time derivative on both sides of Eq.~\refeqn{sigmasaturated2} to get
\begin{subequations}
\begin{empheq}[left ={\dot{\bar{e}}^{b}_{\ast} =\empheqlbrace}]{align}
        \dot{\bar{e}}_{\ast}, & \:\:\:\:\:\:\:\:\text{if } -\rho \|f^{ref}_{\ast}\| \leq \bar{e}_{\ast} \leq \rho \|f^{ref}_{\ast}\|; \label{eqn:sigmasaturateddot2a} \\
        \rho \|\dot{f}^{ref}_{\ast}\|, & \:\:\:\:\:\:\:\:\text{if } \bar{e}_{\ast} > \rho \|f^{ref}_{\ast}\|; \label{eqn:sigmasaturateddot2b} \\
        -\rho \|\dot{f}^{ref}_{\ast}\|, & \:\:\:\:\:\:\:\:\text{if } \bar{e}_{\ast} < -\rho \|f^{ref}_{\ast}\| \label{eqn:sigmasaturateddot2c}.
\end{empheq}
\label{eqn:sigmasaturateddot}
\end{subequations}
The expression of $\dot{\bar{e}}_{\ast}$ for Eq.~\refeqn{sigmasaturateddot2a} could be obtained by taking component-wise first order time derivative on both sides of Eq.~\refeqn{positionerrordynamics} as
\begin{align}
	\dot{\bar{e}}_{\ast} = \ddot{e}_{\ast} + C_{\ast} \dot{e}_{\ast}, \label{eqn:sigmadot1}
\end{align}
where $\dot{e}_{\ast}$, $\ddot{e}_{\ast} \in\Re$ are the first and second order time derivatives of the components of position error $e_{\ast}$ and $C_{\ast} \in\Re$ is the diagonal element of the gain matrix $C$. Substituting $e$ from Eq.~\refeqn{positionerror} to Eq.~\refeqn{sigmadot1} gives
\begin{align}
	\dot{\bar{e}}_{\ast} = \ddot{x}_{\ast} - \ddot{x}^{ref}_{\ast} + C_{\ast} \dot{e}_{\ast}. \label{eqn:sigmadot2}
\end{align}
Substituting Eq.~\refeqn{sigmadot2} to Eq.~\refeqn{sigmasaturateddot2a} gives
\begin{subequations}
\begin{empheq}[left ={\dot{\bar{e}}^{b}_{\ast} =\empheqlbrace}]{align}
\begin{split}
         \ddot{x}_{\ast} - \ddot{x}^{ref}_{\ast} +& C_{\ast} \dot{e}_{\ast}, \\ &\text{if } -\rho \|f^{ref}_{\ast}\| \leq \bar{e}_{\ast} \leq \rho \|f^{ref}_{\ast}\|; \label{eqn:sigmasaturateddot2a2}
\end{split} \\
        \rho \|\dot{f}^{ref}_{\ast}\|, & \:\:\:\:\:\:\:\:\text{if } \bar{e}_{\ast} > \rho \|f^{ref}_{\ast}\|; \label{eqn:sigmasaturateddot2b2} \\
        -\rho \|\dot{f}^{ref}_{\ast}\|, & \:\:\:\:\:\:\:\:\text{if } \bar{e}_{\ast} < -\rho \|f^{ref}_{\ast}\| \label{eqn:sigmasaturateddot2c2}.
\end{empheq}
\label{eqn:sigmasaturateddot2}
\end{subequations}
The first time derivative on both sides of Eq.~\refeqn{force_exerted} gives the impulse:
\begin{align}
\dot{f}_{\ast} = \begin{cases}
	        K^{o}_{\ast}(\dot{x}_{\ast} - \dot{x}^{o}_{\ast}) + D^{o}_{\ast}(\ddot{x}_{\ast} - \ddot{x}^{o}_{\ast}) & \text{in obstacle-space}; \\
		0 & \text{in free-space},
		\end{cases}
\label{eqn:forcedot}
\end{align}
where $K^{o}_{\ast}, D^{o}_{\ast} \in\Re$ are the diagonal elements of $K^{o}$ and $D^{o}$ that represent spring constants and damping coefficients of the surface of the obstacle in $\ast$ direction. $\dot{x}_{\ast}, \dot{x}^{o}_{\ast} \in\Re$ are the linear velocities of the end-effector and the contact point on the obstacle in $\ast$ direction. Likewise, $\ddot{x}_{\ast}, \ddot{x}^{o}_{\ast} \in\Re$ are the linear accelerations of the end-effector and the contact point on the obstacle in $\ast$ direction.

\begin{figure*}[b]
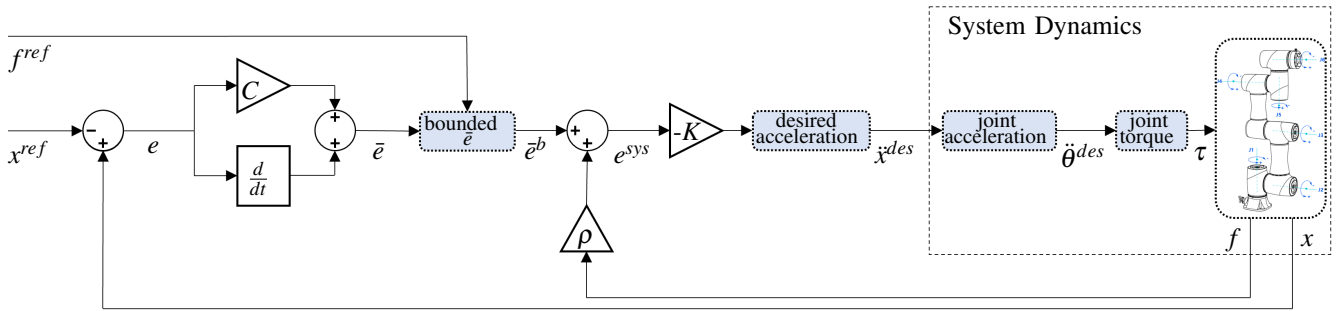

    \centering
    \begin{minipage}[b]{0.995\textwidth}
	\centering
        \begin{overpic}[height=0.23\textwidth]{controller\_architecture}
        	\put(0,11.25){\fontsize{10}{10}\selectfont $x^{ref}$}
		\put(0,18.5){\fontsize{10}{10}\selectfont $f^{ref}$}
		\put(10.5,12){\fontsize{10}{10}\selectfont $e$}
		\put(17.65,16.15){\fontsize{10}{1}\selectfont $C$}
		\put(18,9.55){\fontsize{10}{10}\selectfont $\frac{d}{dt}$}
		\put(27.5,11.55){\fontsize{10}{10}\selectfont $\bar{e}$}
		\put(31.45,13.75){\fontsize{8}{1}\selectfont  $\text{bounded}$}
		\put(34.35,12.65){\fontsize{8}{1}\selectfont $\bar{e}$}
		\put(39,11.55){\fontsize{10}{10}\selectfont $\bar{e}^{b}$}
		\put(43,5.25){\fontsize{10}{10}\selectfont $\rho$}
		\put(45.75,11.25){\fontsize{10}{10}\selectfont $e^{sys}$}
		\put(50.25,13){\fontsize{10}{1}\selectfont $\text{-}K$}
		\put(58,14){\fontsize{8}{1}\selectfont $\text{desired}$}
		\put(56.5,12.75){\fontsize{8}{1}\selectfont $\text{acceleration}$}
		\put(65.5,11.25){\fontsize{10}{10}\selectfont $\ddot{x}^{des}$}
		\put(73.25,14){\fontsize{8}{1}\selectfont $\text{joint}$}
		\put(70.75,12.75){\fontsize{8}{1}\selectfont $\text{acceleration}$}
		\put(79.5,11){\fontsize{10}{10}\selectfont $\ddot{\theta}^{des}$}
		\put(84.75,14){\fontsize{8}{1}\selectfont $\text{joint}$}
		\put(84,12.75){\fontsize{8}{1}\selectfont $\text{torque}$}
		\put(89.55,11.5){\fontsize{10}{10}\selectfont $\tau$}
		\put(92,5){\fontsize{10}{10}\selectfont $f$}
		\put(97.75,5){\fontsize{10}{10}\selectfont $x$}
		\put(71,21){\fontsize{10}{1}\selectfont $\text{System Dynamics}$}
	\end{overpic}
    \end{minipage}

    \caption[Simplified controller architecture]{The simplified architecture of the combined controller showcasing the integration of all the key components along with the dynamics of the 6 DOF manipulator. The blue shaded dotted blocks represent functions that compute the outputs specified by the corresponding label. The input to the manipulator is the joint torque $\tau$ and the output of the manipulator are the force exerted onto the obstacle $f$ and the position of the end-effector $x$.}
    \label{fig:controller}
\end{figure*}

Substituting the expressions for $\dot{\bar{e}}^b{}_{\ast}$ and $\dot{f}_{\ast}$ from Eq.~\refeqn{sigmasaturateddot2} and \refeqn{forcedot} to Eq.~\refeqn{sigmatdynamics3} gives
\begin{align}
\begin{split}
	K_{\ast}{e}^{sys}_{\ast} = - \ddot{x}_{\ast} + \ddot{x}^{ref}_{\ast} &- C_{\ast} \dot{e}_{\ast} - \rho \dot{f}_{\ast}, 	\\ &\text{if } -\rho \|f^{ref}_{\ast}\| \leq \bar{e}_{\ast} \leq \rho \|f^{ref}_{\ast}\|, \label{eqn:totalerror_component_1}
\end{split}
\end{align}
\begin{align}
\begin{split}
	K_{\ast}{e}^{sys}_{\ast} = -K^{o}_{\ast}(\dot{x}_{\ast} - \dot{x}^{o}_{\ast}) -D^{o}_{\ast}(\ddot{x}_{\ast} -& \ddot{x}^{o}_{\ast}) - \rho \|\dot{f}^{ref}_{\ast}\|),    \\ &\text{if } \bar{e}_{\ast} > \rho \|f^{ref}_{\ast}\|,	\label{eqn:totalerror_component_2}
\end{split}
\end{align}
\begin{align}
\begin{split}
	K_{\ast}{e}^{sys}_{\ast} = -K^{o}_{\ast}(\dot{x}_{\ast} - \dot{x}^{o}_{\ast}) -D^{o}_{\ast}(\ddot{x}_{\ast} -& \ddot{x}^{o}_{\ast}) + \rho \|\dot{f}^{ref}_{\ast}\|),     \\ &\text{if } \bar{e}_{\ast} < -\rho \|f^{ref}_{\ast}\|.	\label{eqn:totalerror_component_3}
\end{split}
\end{align}
Substituting $\ddot{x}^{des}_{\ast}$ for $\ddot{x}_{\ast}$ and rearranging Eq.~\refeqn{totalerror_component_1} - \refeqn{totalerror_component_3} gives
\begin{align}
\begin{split}
	\ddot{x}^{des}_{\ast} = -K_{\ast}{e}^{sys}_{\ast} + \ddot{x}^{ref}_{\ast} -& C_{\ast} \dot{e}_{\ast} - \rho \dot{f}_{\ast}, \\ &\text{if } -\rho \|f^{ref}_{\ast}\| \leq \bar{e}_{\ast} \leq \rho \|f^{ref}_{\ast}\|, \label{eqn:desired_acceleration_component_11}
\end{split}
\end{align}
\begin{align}
\begin{split}
	\ddot{x}^{des}_{\ast} = \frac{1}{\rho D^{o}_{\ast}} (-K_{\ast}{e}^{sys}_{\ast} + \rho D^{o}_{\ast} \ddot{x}^{o}_{\ast} -& \rho K^{o}_{\ast} (\dot{x}_{\ast} - \dot{x}^{o}_{\ast}) - \rho \|\dot{f}^{ref}_{\ast}\|), \\ &\text{if } \bar{e}_{\ast} > \rho \|f^{ref}_{\ast}\|,	\label{eqn:desired_acceleration_component_22}
\end{split}
\end{align}
\begin{align}
\begin{split}
	\ddot{x}^{des}_{\ast} = \frac{1}{\rho D^{o}_{\ast}} (-K_{\ast}{e}^{sys}_{\ast} + \rho D^{o}_{\ast} \ddot{x}^{o}_{\ast} -& \rho K^{o}_{\ast} (\dot{x}_{\ast} - \dot{x}^{o}_{\ast}) + \rho \|\dot{f}^{ref}_{\ast}\|), \\ &\text{if } \bar{e}_{\ast} < -\rho \|f^{ref}_{\ast}\|.	\label{eqn:desired_acceleration_component_33}
\end{split}
\end{align}

The next task is to compute the joint torque $\tau$ such that the resulting linear end-effector acceleration is $\ddot{x}^{des}$. The terms joint space and task spaces are used to describe motion in terms of joint angles and Cartesian pose, respectively. The task space acceleration $\ddot{\mathscr{X}} \in\Re^6$ of the end-effector frame, i.e., frame $\{7\}$, is expressed as
\begin{align}
	\ddot{\mathscr{X}} = \begin{bmatrix}\ddot{x} \\ \alpha \end{bmatrix}, \label{eqn:taskspaceaccel}
\end{align}
where $\ddot{x}$, $\alpha \in\Re^3$ are the linear and angular accelerations of the end-effector frame expressed with respect to frame $\{0\}$. The relationship between the joint space acceleration $\ddot{\theta}$ and the task space acceleration $\ddot{\mathscr{X}}$ is
\begin{align}
	\ddot{\theta} = J^{-1}_{7}(\ddot{\mathscr{X}} - \dot{J}_{7}\dot{\theta}), \label{eqn:tasktojointaccel}
\end{align}
where $\dot{\theta} \in\Re^6$ is the joint velocity of the 6 DOF manipulator, $J_{7} \in\Re^{6 \times 6}$ is the Jacobian expressed with respect to frame $\{0\}$ that maps the joint velocity to the end-effector velocity, and $\dot{J}_{7}$ is the first time derivative of $J_{7}$.

One can substitute $\ddot{x}^{des}$ for $\ddot{x}$ and select an arbitrary $\alpha$ to compute the desired joint acceleration $\ddot{\theta}^{des}$ using Eq.~\refeqn{tasktojointaccel}. The input joint torque $\tau$ to the 6 DOF manipulator is thus obtained from the Eq.~\refeqn{generaldynamics} by substituting $\ddot{\theta}^{des}$ for $\ddot{\theta}$. A simplified architecture of the controller with the key components is shown in Fig.~\ref{fig:controller}.

\subsection{Controller Gains}

The controller gains $C$ and $K$ were first mentioned when the first order dynamics of the position error $e$ and total error $e^{sys}$ were introduced in Eq.~\refeqn{positionerrordynamics} and \refeqn{sigma_t_dynamics}, respectively. The analysis of the influence of $C$ and $K$ on the response of the manipulator helps in tuning the controller accordingly to get the desired response. The effects of the controller gains are straightforward: the magnitudes of the diagonal elements of $C$ and $K$ dictate the rate of stability of the position error $e$ and the total error $e^{sys}$, respectively. In other words, $C$ and $K$ dictate how quickly $e$ and $e^{sys}$ approach $\mathbf{0}$, respectively.

From Eq.~\refeqn{positionerrordynamics} and \refeqn{positionerrordynamicssoln1}, it is observed that the larger values of the diagonals of $C$ stabilize the initial position error $e(0) \in\Re^3$ quicker; this results in the end-effector to follow the trajectory at faster speeds. Likewise, similar analysis of Eq.~\refeqn{sigma_t_dynamics} suggests that larger values of the diagonals of $K$ stabilize the total error $e^{sys}$ quicker. As a result, the end-effector follows the trajectory closely and the force exerted by the end-effector gets bounded by the reference force faster.

\section{SIMULATION RESULTS \& DISCUSSION}
\subsection{Numerical Simulations}

\begin{figure}[h]
	\centerline{
		\hspace*{0.00001\columnwidth}
		\subfigure{\includegraphics[width=0.31\columnwidth]{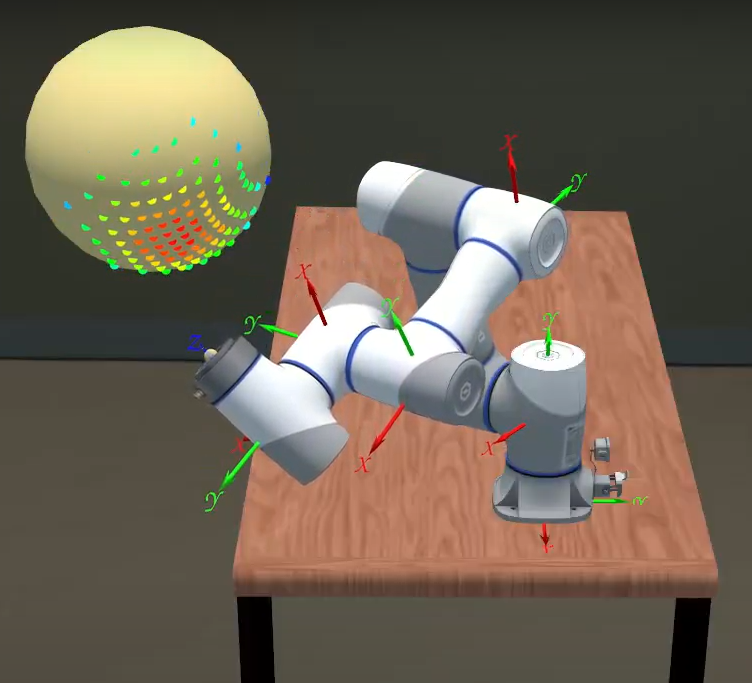}}
		\hspace*{0.00001\columnwidth}
		\subfigure{\includegraphics[width=0.31\columnwidth]{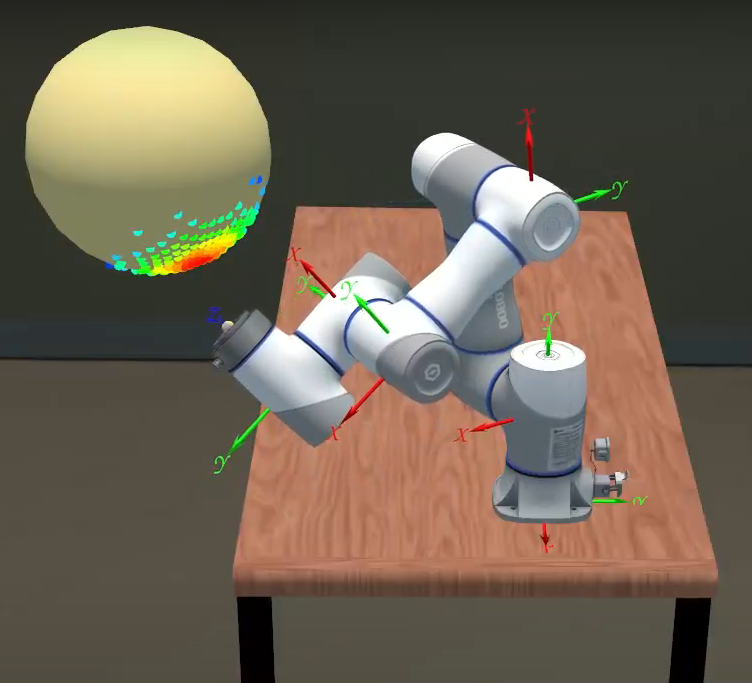}}
		\hspace*{0.00001\columnwidth}
		\subfigure{\includegraphics[width=0.31\columnwidth]{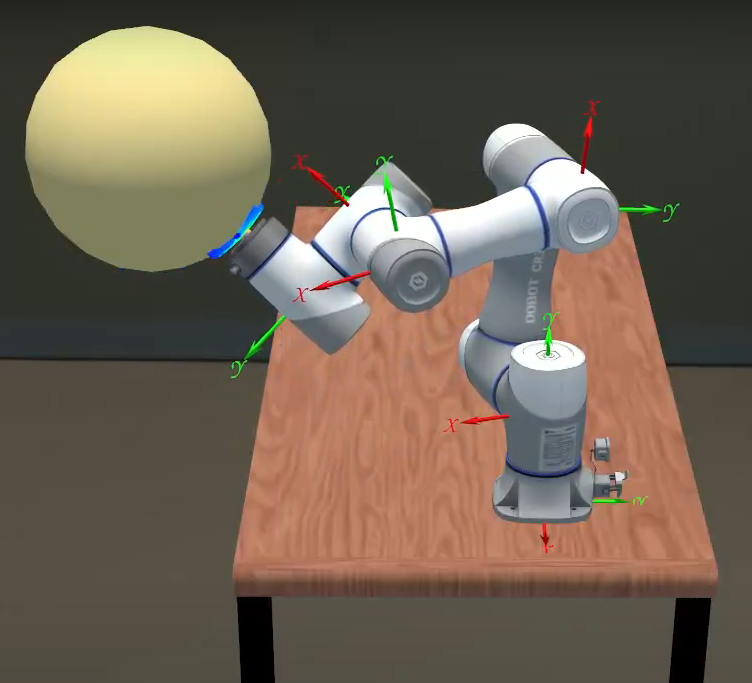}}
	}
	\caption[VxSIM\texttrademark{} force sensor]{From the left to the right are the snapshots of the interface of VxSIM\texttrademark{} showing the obstacle detection feature while the end-effector moves from the free space to the obstacle space. The colored dots represent the locations where the rays emitted by the obstacle detector intersect the obstacle. Red and blue dots represent the closest and the farthest intersection points, respectively. Other intersection points are color mapped accordingly.}\label{fig:forcedetector}
\end{figure}

Two numerical examples are created to verify that the combined controller meets both the motion and force control objectives. VxSIM\texttrademark{}, an advanced software simulation framework for robotic applications, is used to instantiate a model of the 6-DOF robotic manipulator as a digital twin.  The twin is equipped with a force sensor at the end-effector as shown in Fig.~\ref{fig:forcedetector}; a small sphere at the end-effector representing the contact surface of the force sensor.  Any object that intersects or collides with sphere surface is considered to be in contact with the robot. The force sensor utilizes the VxSIM\texttrademark{} detector process module to determine intersections between objects, utilizing an advanced ray trace methodology to compute the distance from the end-effector to an obstacle and identify collisions. Upon contact, i.e., when an object’s geometry intersects with the small sphere, the penetration distance is computed then utilized by the force sensor module to compute the force exerted, applied between the obstacle and end-effector (i.e., action-reaction force) as the real systems undergoes as well.


This model is then used to conduct two numerical simulations to study the effects of the contact forces with the force controller.  For both the simulations, the controller gains $C_x$, $C_y$, and $C_z$ are $150 \;s^{-1}$ and $K_x$, $K_y$, and $K_z$ are $350 \;s^{-1}$. Similarly, the force modeling parameters, spring constants $K^{o}_x$, $K^{o}_y$, and $K^{o}_z$ are $250 \times 10^3 \;N/m$ and the damping coefficients $D^{o}_x$, $D^{o}_y$, and $D^{o}_z$ are $100 \;Ns/m$. The positive constant $\rho$ is selected as $0.05 \;s/kg$. The desired acceleration $\ddot{x}^{des}$ and the input torque $\tau$ are saturated to $\pm 50 \; m/s^{2}$ and $\pm 250 \;Nm$, respectively. Also, the reference force for both the simulations is
\begin{align}
	f^{ref} = \begin{bmatrix}
				40 \\50 \\60
			\end{bmatrix}.
	\label{eqn:referenceforce}
\end{align}

For the first simulation, the initial configuration of the manipulator $\theta_0 = [ 200^\circ,\;50^\circ,\;40^\circ,\;30^\circ,\;-10^\circ,\;-20^\circ ]^{\T}$. A spherical obstacle of radius $0.125\; m$ with its center fixed at $[ 0.35\; m,\;0.2\; m,\;0.125\; m ]^{\T}$ is placed in the scene. The reference trajectory in the task space is
\begin{align}
	x^{ref} = \begin{bmatrix}
				0.42 + (0.3 - 0.42) \sin{(\frac{\pi}{40}t)} \\[3pt]
				0.3994 - 0.3994 \sin{(\frac{\pi}{40}t)} \\[3pt]
				0.15 - \frac{1}{5} \sin({\frac{4 \pi}{20} t})
			\end{bmatrix}.
	\label{eqn:referencetrajectory1}
\end{align}

For the second simulation, the initial configuration of the manipulator $\theta_0 = [ -48.3^\circ,\;-8^\circ,\;-45^\circ,\;-71^\circ,\;-54^\circ,\;0^\circ ]^{\T}$. A spherical obstacle of radius $0.15\; m$ with its center fixed at $[ -0.075\; m,\;-0.5\; m,\;0.45\; m ]^{\T}$ is placed in the scene. The reference trajectory in the task space is
\begin{align}
	x^{ref} = \begin{bmatrix}
				0.1 + 0.15 \sin{(\frac{\pi}{10}t)} \\[3pt]
				-0.4 \\[3pt]
				0.4 + 0.15 \cos{(\frac{\pi}{10}t)}
			\end{bmatrix}.
	\label{eqn:referencetrajectory2}
\end{align}

%

\begin{figure}[h]
    \centering

    \begin{minipage}[b]{0.9\columnwidth}
	\centering
        \begin{overpic}[height=0.7\columnwidth]{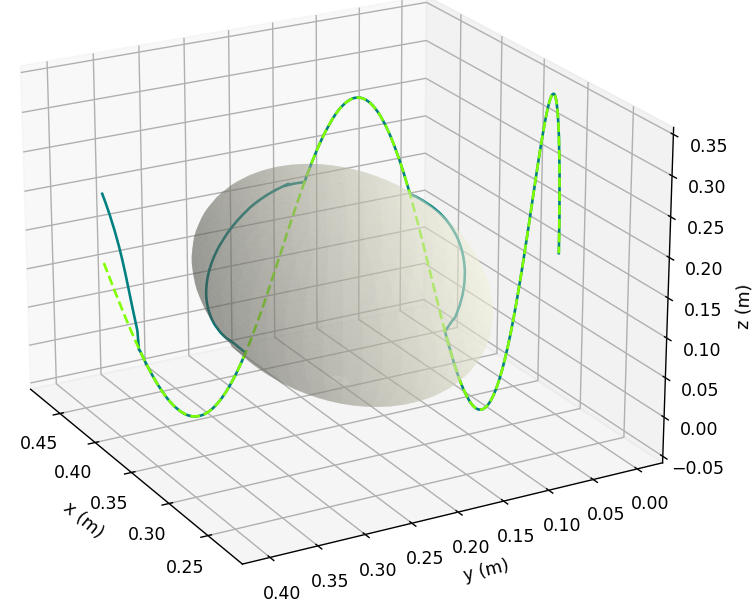}
        \end{overpic}
    \end{minipage}
	\caption[Path traced by end-effector (Wavy path)]{Simulation 1 (wavy path): The path of the end-effector (solid blue) tracks the reference path (dashed green) in the free space and goes around the boundary of the obstacle (gray sphere) when the reference path is inside the obstacle.}
    \label{fig:wavypath1}
\end{figure}

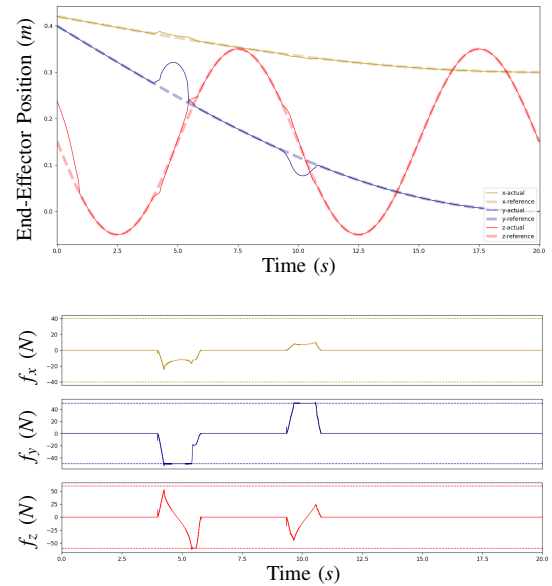
\begin{figure}[h]
    \begin{minipage}[b]{0.99\columnwidth}
	\centering
        \begin{overpic}[width=0.8\columnwidth]{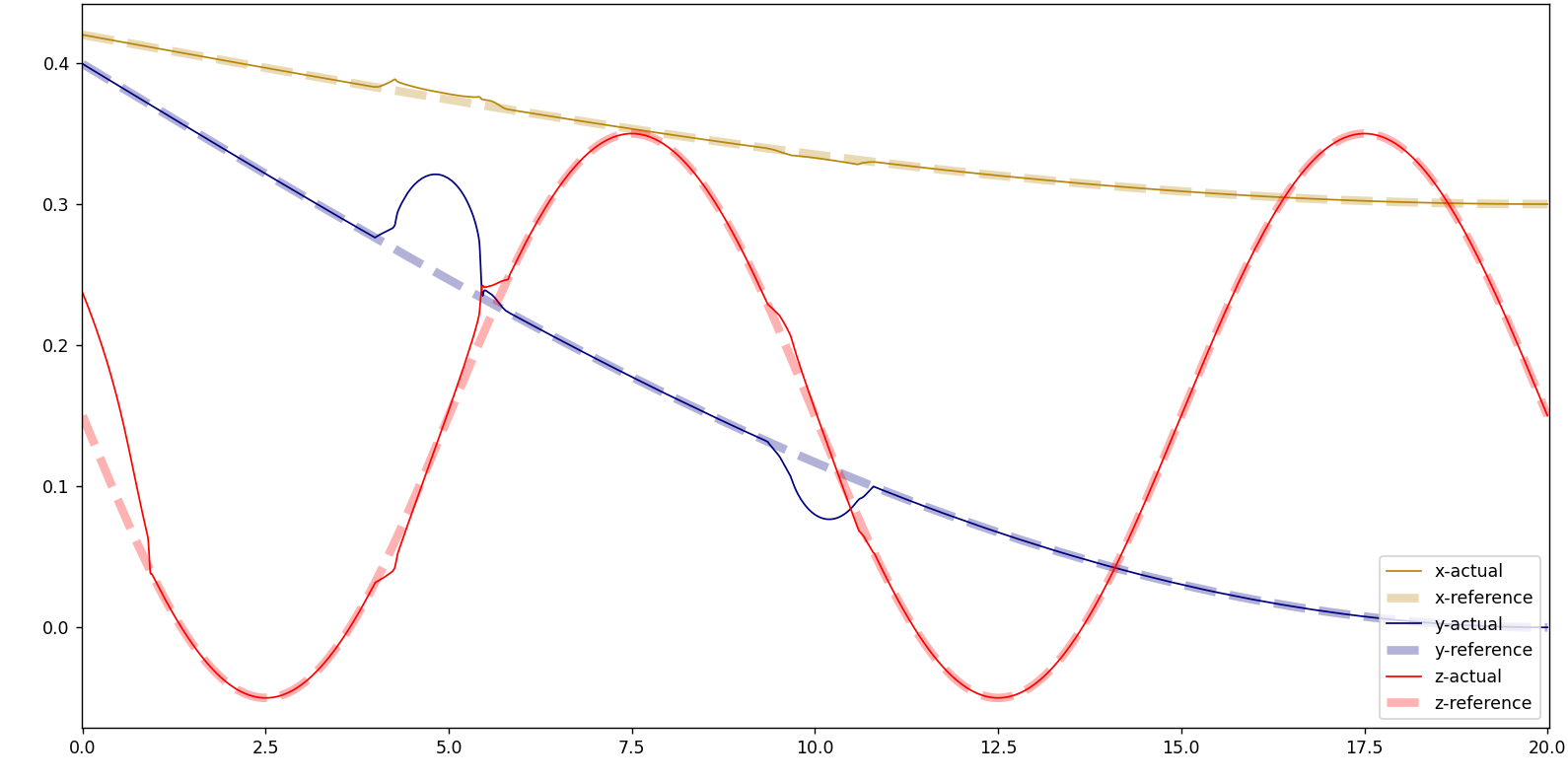}
		\put(-2.75,4){\fontsize{8}{2}\selectfont \rotatebox{90}{End-Effector Position ($m$)}}
		\put(45,-2){\fontsize{8}{2}\selectfont {Time ($s$)}}
        \end{overpic}
\vspace*{0.75cm}
    \end{minipage}
    \begin{minipage}[b]{0.99\columnwidth}
	\centering
        \begin{overpic}[width=0.8\columnwidth]{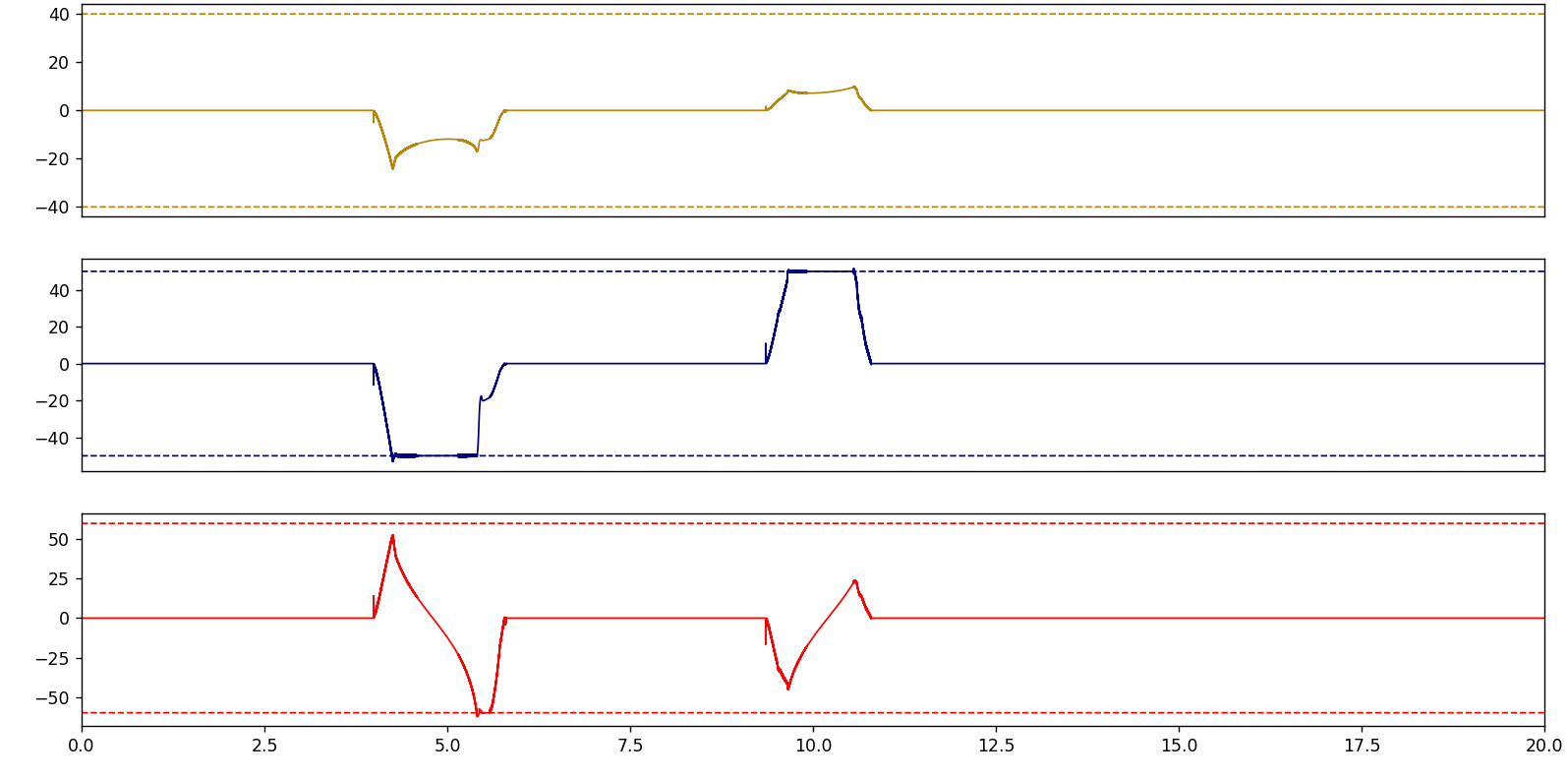}
		\put(-3.25,36){\fontsize{8}{2}\selectfont \rotatebox{90}{$f_{x}$ ($N$)}}
		\put(-3.25,20.5){\fontsize{8}{2}\selectfont \rotatebox{90}{$f_{y}$ ($N$)}}
		\put(-3.25,4){\fontsize{8}{2}\selectfont \rotatebox{90}{$f_{z}$ ($N$)}}
		\put(45,-2){\fontsize{8}{2}\selectfont {Time ($s$)}}
        \end{overpic}
    \end{minipage}

    \caption[End-effector trajectories, \& forces (Wavy path)]{End-effector trajectories, \& forces for Simulation 1 (wavy path): Top: The trajectories of the end-effector (solid) with their reference values (dashed). The trajectories and the corresponding references in $x$, $y$, and $z$ directions are colored as yellow, blue, and red respectively. Bottom: The forces exerted by the end-effector (solid) along with the positive and the negative of the magnitude of the corresponding reference values (dashed) in $x$, $y$, and $z$ directions as a function of time.}
    \label{fig:wavypath2}
\end{figure}

\begin{figure}[h]
    \centering

    \begin{minipage}[b]{0.9\columnwidth}
	\centering
        \begin{overpic}[height=0.7\columnwidth]{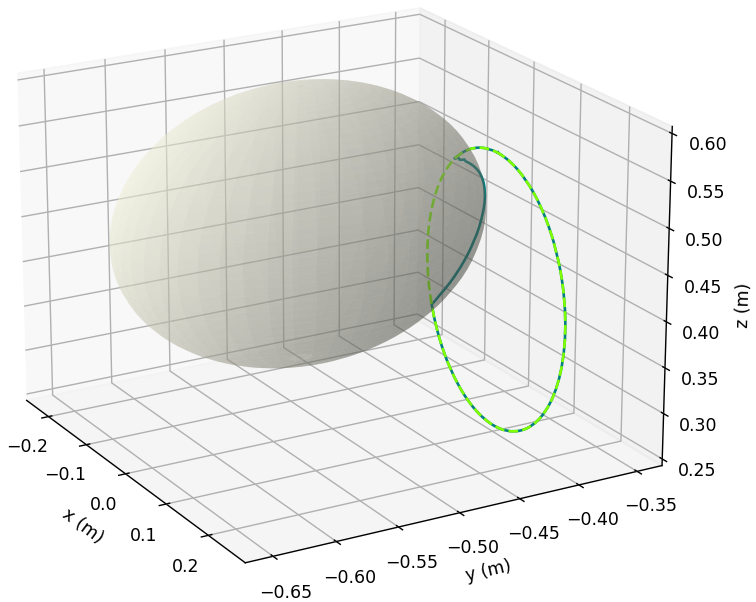}
        \end{overpic}
    \end{minipage}
	\caption[Path traced by end-effector (Circular path)]{Simulation 2 (circular path): The path of the end-effector (solid blue) tracks the reference path (dashed green) in the free space and goes around the boundary of the obstacle (gray sphere) when the reference path is inside the obstacle.}
	\label{fig:circularpath1}
\end{figure}

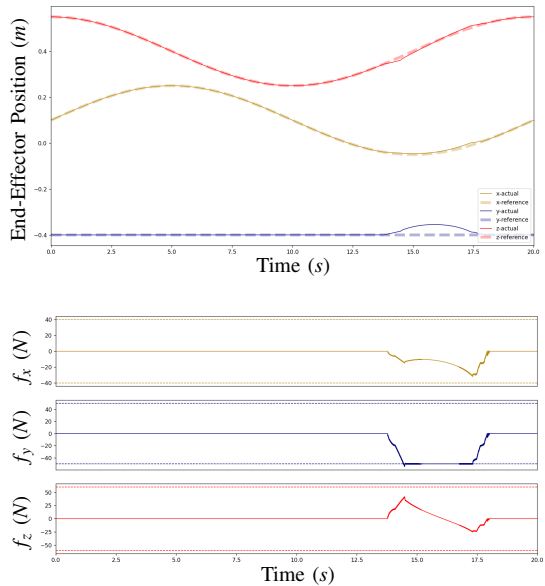
\begin{figure}[h]
    \begin{minipage}[b]{0.99\columnwidth}
	\centering
        \begin{overpic}[width=0.8\columnwidth]{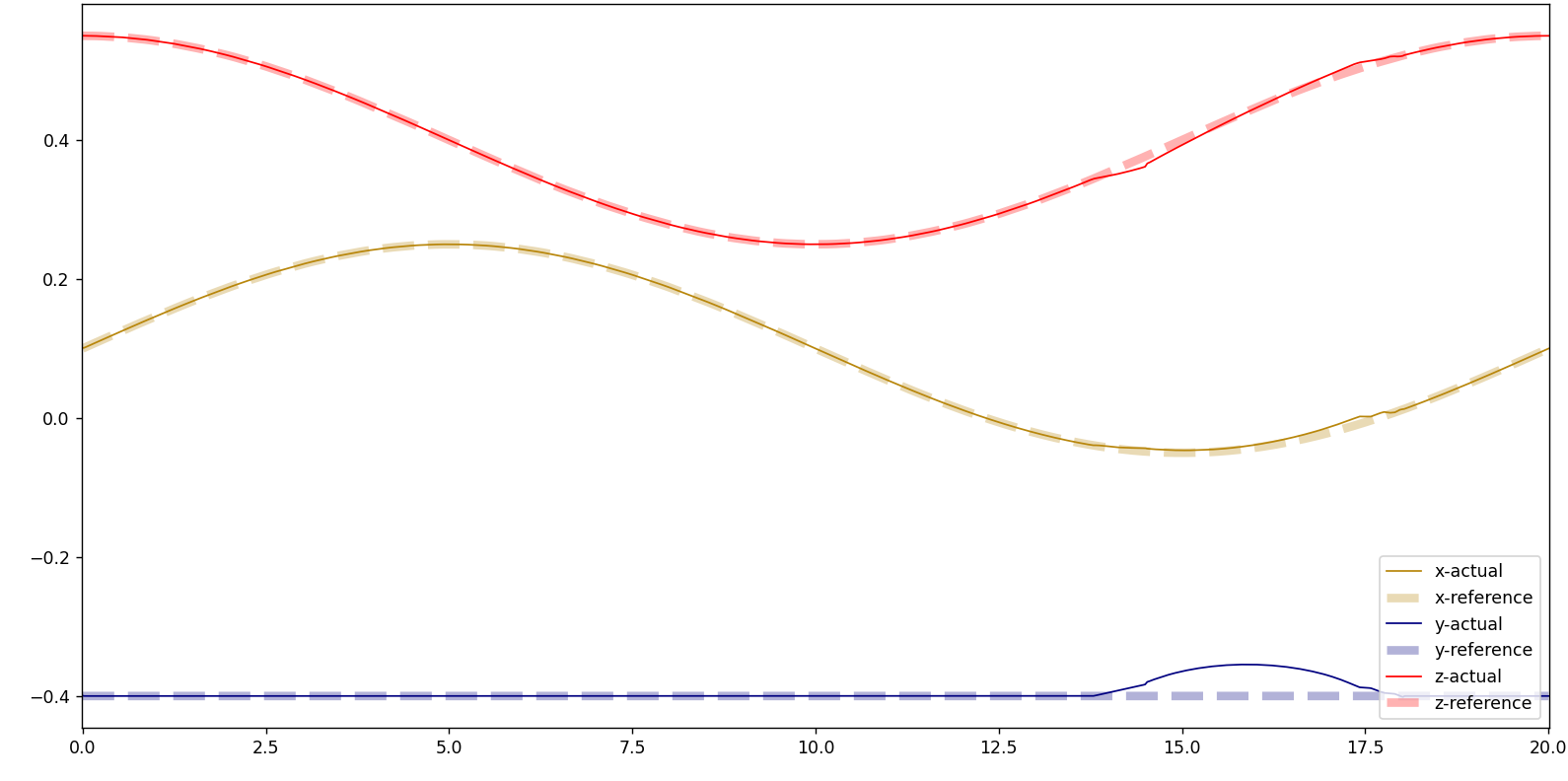}
		\put(-2.75,4){\fontsize{8}{2}\selectfont \rotatebox{90}{End-Effector Position ($m$)}}
		\put(45,-2){\fontsize{8}{2}\selectfont {Time ($s$)}}
        \end{overpic}
\vspace*{0.75cm}
    \end{minipage}
    \begin{minipage}[b]{0.99\columnwidth}
	\centering
        \begin{overpic}[width=0.8\columnwidth]{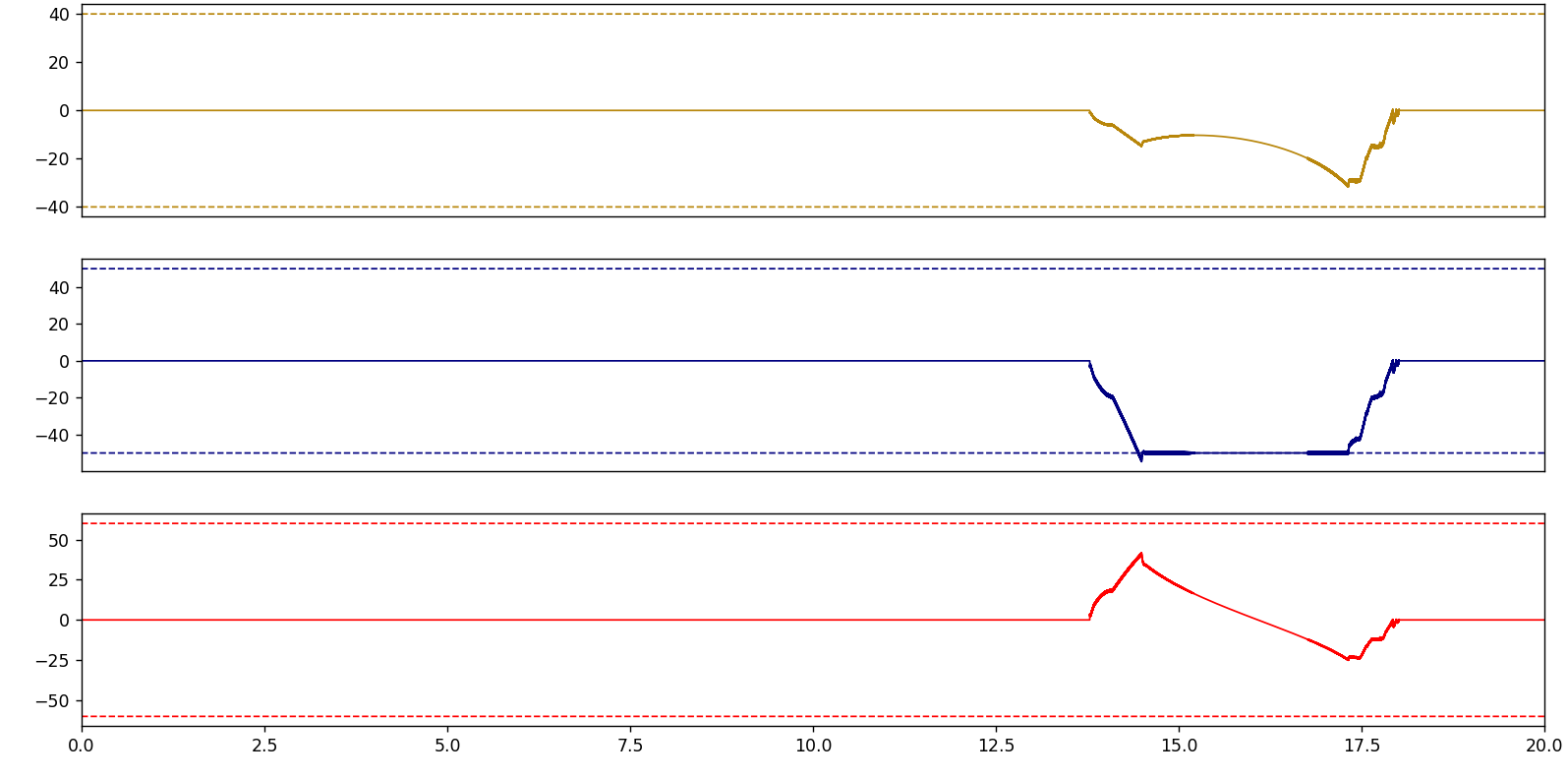}
		\put(-3.25,36){\fontsize{8}{2}\selectfont \rotatebox{90}{$f_{x}$ ($N$)}}
		\put(-3.25,20.5){\fontsize{8}{2}\selectfont \rotatebox{90}{$f_{y}$ ($N$)}}
		\put(-3.25,4){\fontsize{8}{2}\selectfont \rotatebox{90}{$f_{z}$ ($N$)}}
		\put(45,-2){\fontsize{8}{2}\selectfont {Time ($s$)}}
        \end{overpic}
    \end{minipage}

	\caption[End-effector trajectories, \& forces (Circular path)]{Simulation 2 (circular path): Top: The trajectories of the end-effector (solid) with their reference values (dashed). The trajectories and the corresponding references in $x$, $y$, and $z$ directions are colored as yellow, blue, and red respectively. Bottom: The forces exerted by the end-effector (solid) along with the positive and the negative of the magnitude of the corresponding reference values (dashed) in $x$, $y$, and $z$ directions as a function of time.}
	\label{fig:circularpath2}
\end{figure}

The path followed by the end-effector along with the trajectories and the forces it exerted onto the obstacle upon contact for the first simulation are shown in Fig.~\ref{fig:wavypath1} and~\ref{fig:wavypath2} and for the second simulation are shown in Fig.~\ref{fig:circularpath1} and~\ref{fig:circularpath2}.

\subsection{Discussion}

 According to Fig.~\ref{fig:wavypath1} and \ref{fig:circularpath1}, both numerical examples show that the end-effector follows the reference path when the end-effector is in free space and goes around the boundary of the obstacle when the end-effector encounters an obstacle. The position-time plots of the end-effector, i.e., the top plots of Fig.~\ref{fig:wavypath2} and \ref{fig:circularpath2} show that the end-effector performs trajectory tracking when the end-effector is in free space since the end-effector trajectories are aligned with the reference trajectories. The end-effector trajectories deviate from the reference trajectories when the obstacle blocks the reference trajectories. Also, note that in the simulation presented in Fig.~\ref{fig:wavypath1} and \ref{fig:wavypath2}, the initial location of the end-effector is different than the starting location of the reference path. So, for approximately $1$ second, the trajectory of the end-effector along $z$-axis is not overlapped with the reference trajectory even though the reference trajectory is not blocked by the obstacle.

The bottom plots of Fig.~\ref{fig:wavypath2} and \ref{fig:circularpath2} show that the forces that the end-effector exerts onto the obstacle is bounded by the reference forces which are specified by the user. Note that the force that the end-effector exerts when the end-effector is in free space is $0$; the forces arise when there is contact. Therefore, the results verify that the end-effector follows the reference trajectory when in free space and limits the amount of force exerted onto the obstacle when it comes in contact with the obstacle.

\section{CONCLUSIONS}

The paper presents a controller design for motion and force control for a 6 DOF serial manipulator. The dynamics of the manipulator is analyzed using Lagrangian method. The contact between the end-effector and the obstacle is modeled as a spring-damper system. A scheme for trajectory tracking is discussed and then a modification to the motion control scheme is performed to incorporate the force control strategy. Therefore, the presented motion-force controller performs trajectory tracking when the end-effector is in free space and restricts the force exerted onto the obstacle when the end-effector makes a contact with an obstacle. The effects of the controller gains on the response of the manipulator are discussed which is useful while tuning them. The simulations verify that the proposed combined controller performs both motion and force control.



\addtolength{\textheight}{-12cm}   


\end{document}